\documentclass[10pt,twocolumn,letterpaper]{article}

\usepackage[table,dvipsnames]{xcolor}
\usepackage{tikz} 

\usepackage{cvpr}
\usepackage{times}
\usepackage{epsfig}
\usepackage{graphicx}
\usepackage{amsmath}
\usepackage{amssymb}
\usepackage{paralist}

\usepackage{booktabs} 
\usepackage{etoolbox,siunitx} 

\definecolor{olivegreen}{RGB}{0,170,0}
\definecolor{darkred}{RGB}{220,100,10}
\definecolor{tealblue}{RGB}{20,100,200}

\newcommand{\osvos}{caelles2017one}
\newcommand{\onavos}{voigtlaender17BMVC}
\newcommand{\ctn}{Jang2017}
\newcommand{\sfls}{Cheng2017}
\newcommand{\vpn}{Jampani2017}
\newcommand{\msk}{Perazzi2017}
\newcommand{\ofl}{Tsai2016}
\newcommand{\bvs}{NicolasMaerki2016}
\newcommand{\fcp}{Perazzi2015}
\newcommand{\jmp}{Fan2015}
\newcommand{\hvs}{Grundmann2010}
\newcommand{\sea}{Ramakanth2014}

\newcommand{\npml}{PML (Ours)}
\newcommand{\nosvos}{OSVOS}
\newcommand{\nonavos}{OnAVOS}
\newcommand{\nmsk}{MSK}

\newcommand{\nsfls}{SFL}
\newcommand{\nctn}{CTN}
\newcommand{\nvpn}{VPN}
\newcommand{\nofl}{OFL}
\newcommand{\nbvs}{BVS}
\newcommand{\nfcp}{FCP}
\newcommand{\njmp}{JMP}
\newcommand{\nhvs}{HVS}
\newcommand{\nsea}{SEA}

\newcommand{\J}{\mathcal{J}}
\newcommand{\F}{\mathcal{F}}
\newcommand{\T}{\mathcal{T}}

\definecolor{rowblue}{RGB}{220,230,240}

\usepackage{pgfplots} 
\usepackage{pgfplotstable} 
\pgfplotsset{compat=newest} 
\usetikzlibrary{plotmarks} 
\usetikzlibrary{colorbrewer} 
\usepackage{subfiles}

\cvprfinalcopy 

\def\cvprPaperID{1550} 

\ifcvprfinal\fi
\begin{document}

\title{Blazingly Fast Video Object Segmentation with Pixel-Wise Metric Learning}

\author{Yuhua Chen$^1$\hspace{10mm}Jordi Pont-Tuset$^1$\hspace{10mm}Alberto  Montes$^1$\hspace{10mm}Luc Van Gool$^{1,2}$\\[2mm]
$^1$Computer Vision Lab, ETH Zurich\hspace{10mm}
$^2$VISICS, ESAT/PSI, KU Leuven\\[-1.5pt]
{\tt\small \{yuhua.chen,jponttuset,vangool\}@vision.ee.ethz.ch,  malberto@student.ethz.ch}
}
\maketitle
\vspace{-3mm}

\begin{abstract}
This paper tackles the problem of video object segmentation,  given some user annotation which indicates the object of interest. The problem is formulated as pixel-wise retrieval in a learned embedding space: we embed pixels of the same object instance into the vicinity of each other, using a fully convolutional network trained by a modified triplet loss as the embedding model. Then the annotated pixels are set as reference and the rest of the pixels are classified using a nearest-neighbor approach. The proposed method supports different kinds of user input such as segmentation mask in the first frame (semi-supervised scenario), or a sparse set of clicked points (interactive scenario). In the semi-supervised scenario, we achieve results competitive with the state of the art but at a fraction of computation cost (275 milliseconds per frame). In the interactive scenario where the user is able to refine their input iteratively, the proposed method provides instant response to each input, and reaches comparable quality to competing methods with much less interaction.
\end{abstract}

\vspace{-2mm}
\section{Introduction}
Immeasurable amount of multimedia data is recorded and shared in the current era of the Internet. Among it, video is one of the most common and rich modalities, albeit it is also one of the most expensive to process. Algorithms for fast and accurate video processing thus become crucially important for real-world applications. Video object segmentation, i.e.\ classifying the set of pixels of a video sequence into the object(s) of interest and background, is among the tasks that despite having numerous and attractive applications, cannot currently be performed in a satisfactory quality level and at an acceptable speed. The main objective of this paper is to fill in this gap: we perform video object segmentation at the accuracy level comparable to the state of the art while keeping the processing time at a speed that even allows for real-time human interaction.

Towards this goal, we model the problem in a simple and intuitive, yet powerful and unexplored way: we formulate video object segmentation as pixel-wise retrieval in a learned embedding space. Ideally, in the embedding space, pixels belonging to the same object instance are close together and pixels from other objects are further apart. We build such embedding space by learning a Fully Convolutional Network (FCN) as the embedding model, using a modified triplet loss tailored for video object segmentation, where no clear correspondence between pixels is given. Once the embedding model is learned, the inference at test-time only needs to compute the embedding vectors with a forward pass for each frame, and then perform a per-pixel nearest neighbor search in the embedding space to find the most similar annotated pixel. The object, defined by the user annotation, can therefore be segmented throughout the video sequence.

\begin{figure}[t]
\includegraphics[width=\linewidth]{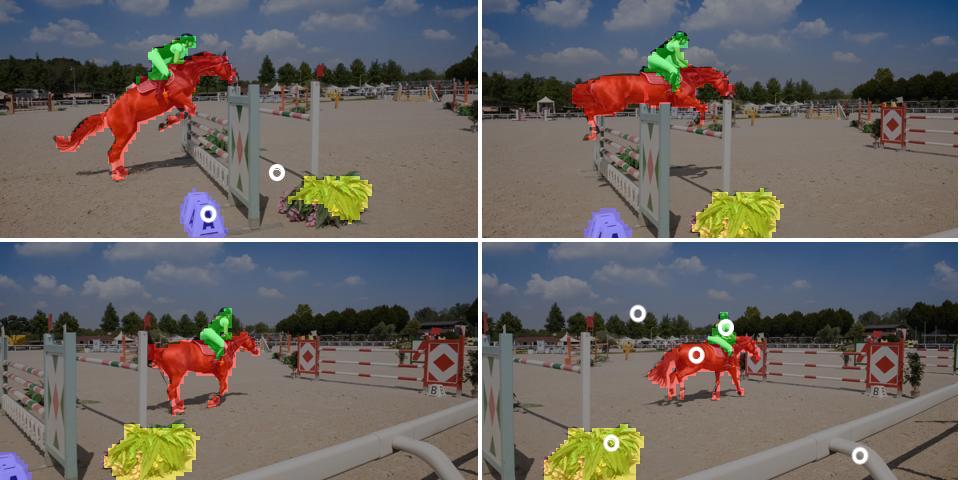}
\vspace{-3mm}
\caption{\textbf{Interactive segmentation using our method:} The white circles represent the clicks where the user has provided an annotation, the colored masks show the resulting segmentation in a subset of the sequence's frames. }
\label{fig:cover}
\vspace{-2mm}
\end{figure}

There are several main advantages of our formulation: Firstly, the proposed method is highly efficient as there is no fine-tuning in test time, and it only requires a single forward pass through the embedding network and a nearest-neighbor search to process each frame. Secondly, our method provides the flexibility to support different types of user input (i.e. clicked points, scribbles, segmentation masks, etc.) in an unified framework. Moreover, the embedding process is independent of user input, thus the embedding vectors do not need to be recomputed when the user input changes, which makes our method ideal for the interactive scenario. We show an example in Figure~\ref{fig:cover}, where the user aims to segment several objects in the video: The user can iteratively refine the segmentation result by gradually adding more clicks on the video, and get feedback immediately after each click.

The proposed method is evaluated on the DAVIS 2016~\cite{Perazzi2016} and DAVIS 2017~\cite{Pont-Tuset_arXiv_2017} datasets, both in the semi-supervised and interactive scenario. In the context of semi-supervised Video Object Segmentation (VOS), where the full annotated mask in the first frame is provided as input, we show that our algorithm presents the best trade-off between speed and accuracy, with 275 milliseconds per frame and $\J\&\F$=77.5\% on DAVIS 2016. In contrast, better performing algorithms start at 8 seconds per frame, and similarly fast algorithms reach only 60\% accuracy. Where our algorithm shines best is in the field of interactive segmentation, with only 10 clicks on the whole video we can reach an outstanding 74.5\% accuracy.

\section{Related Work}

\paragraph{Semi-Supervised and Unsupervised Video Object Segmentation:}\rule{0mm}{1mm}\\
The aim of video object segmentation is to segment a specific object throughout an input video sequence. Driven by the surge of deep learning, many approaches have been developed and performance has improved dramatically. Dependent on the amount of supervision, methods can be roughly categorized into two groups: semi-supervised and unsupervised. 

Semi-supervised video object segmentation methods take the segmentation mask in the first frame as input. MaskTrack~\cite{Perazzi2017} propagates the segmentation from the previous frame to the current one, with optical flow as input. OSVOS~\cite{caelles2017one} learns the appearance of the first frame by a FCN, and then segments the remaining frames in parallel. Follow-up works extend the idea with various techniques, such as online adaptation~\cite{voigtlaender17BMVC}, semantic instance segmentation~\cite{caelles2017semantically,maninis2017video}. Other recent techniques obtain segmentation and flow simultaneously~\cite{Cheng2017,Tsai2016}, train a trident network to improve upon the errors of optical flow propagation~\cite{Jang2017}, or use a CNN in the bilateral space~\cite{Jampani2017}.

Unsupervised video object segmentation, on the other hand, uses only video as input. These methods typically aim to segment the most salient object from cues such as motion and appearance. The current leading technique~\cite{koh2017primary} use region augmentation and reduction to refine object proposals to estimate the primary object in a video. \cite{jain2017fusionseg} proposes to combine motion and appearance cues with a two-stream network. Similarly, \cite{tokmakov2017learning} learns a two-stream network to encode spatial and temporal features, and a memory module to capture the evolution over time.

In this work, we focus on improving the efficiency of video object segmentation to make it suitable for real-world applications where rapid inference is needed.
We do so by, in contrast to previous techniques using deep learning, not performing test-time network fine-tuning and not relying on optical flow or previous frames as input.

\paragraph{Interactive Video Object Segmentation:}\rule{0mm}{1mm}\\
Interactive Video Object Segmentation relies on iterative user interaction to segment the object of interest. Many techniques have been proposed for the task. Video Cutout~\cite{Wang2005} solves a min-cut labeling problem over a hierarchical mean-shift segmentation of the set of video frames, from user-generated foreground and background scribbles. The pre-processing plus post-processing time is in the order of an hour, while the time between interactions is in the order of tens of seconds. A more local strategy is LIVEcut~\cite{Price2009}, where the user iteratively corrects the propagated mask frame to frame and the algorithm learns from it. The interaction response time is reduced significantly (seconds per interaction), but the overall processing time is comparable. TouchCut~\cite{Wang2014} simplifies the interaction to a single point in the first frame, and then propagates the results using optical flow. \textit{Click carving}~\cite{Jain2016} uses point clicks on the boundary of the objects to fit object proposals to them. \textit{A few strokes}~\cite{Nagaraja2015} are used to segment videos based on point trajectories, where the interaction time is around tens of seconds per video. A click-and-drag technique~\cite{Pont-Tuset2015a} is used to label per-frame regions in a hierarchy and then propagated and corrected. 

In contrast to most previous approaches, our method response time is almost immediate, and the pre-processing time is 275 milliseconds per frame, making it suitable to real-world use.

\paragraph{Deep Metric Learning:}\rule{0mm}{1mm}\\
Metric learning is a classical topic and has been widely studied in the learning community~\cite{weinberger2009distance,chechik2010large}. Following the recent success of deep learning, deep metric learning has gained increasing popularity~\cite{sohn2016improved}, and has become the cornerstone of many computer vision tasks such as person re-identification~\cite{cheng2016person,xiao2016learning}, face recognition~\cite{schroff2015facenet}, or unsupervised representation learning~\cite{wang2015unsupervised}. The key idea of deep metric learning is usually to transform the raw features by a network and then compare the samples in the embedding space directly. Usually metric learning is performed to learn the similarity between images  or patches, and methods based on pixel-wise metric learning are limited. Recently, \cite{fathi2017semantic} exploits metric learning at the pixel level for the task of instance segmentation.

In this work, we learn an embedding where pixels of the same instance are aimed to be close to each other, and we formulate video object segmentation as a pixel-wise retrieval problem. The formulation is inspired also by works in image retrieval~\cite{smeulders2000content,qin2014learning}.

\begin{figure*}[h]
\centering
\includegraphics[width=0.8\linewidth]{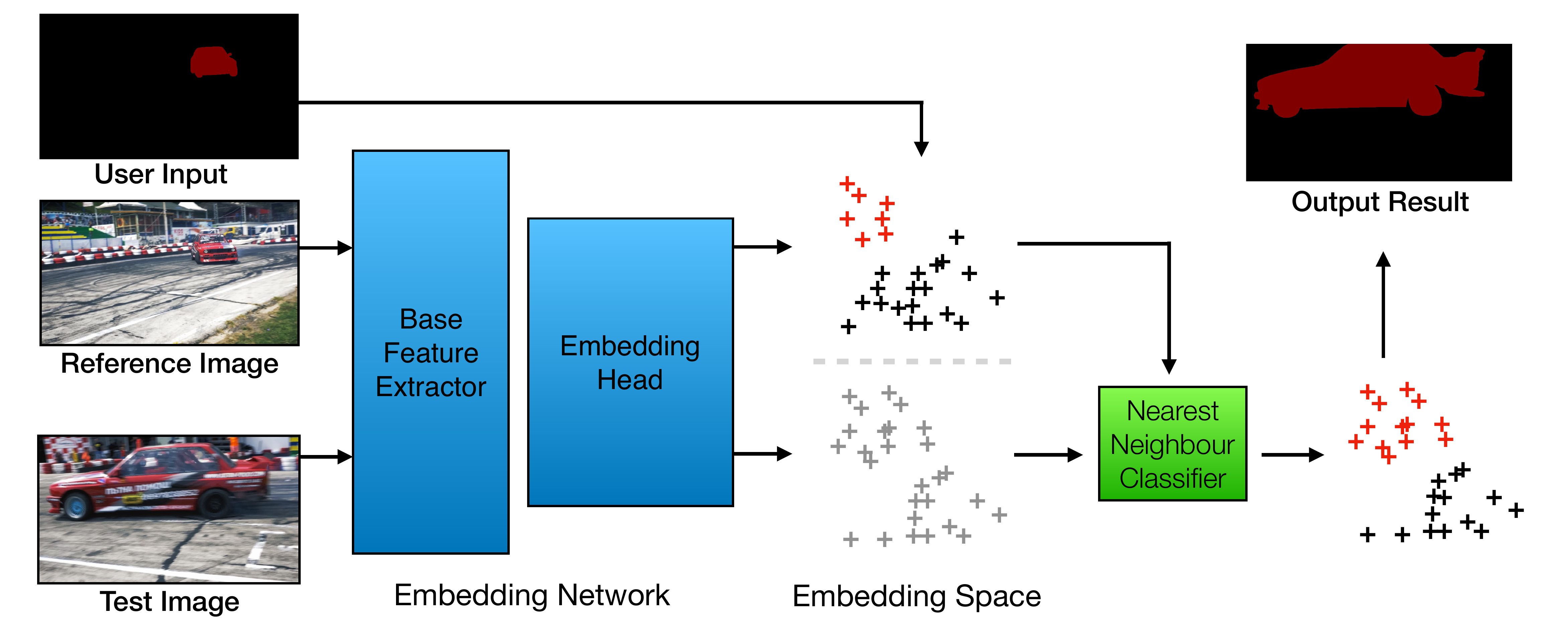}
\caption{\textbf{Overview of the proposed approach:} Here we assume the user input is provided in the form of full segmentation mask for the reference frame, but interactions of other kind are supported as well. }
\label{fig:architecture}
\end{figure*}

\vspace{-2mm}

\section{Proposed Method}
\subsection{Overview}
\label{sec:approaches}

In this work, we formulate video object segmentation as a pixel-wise retrieval problem, that is, for each pixel in the video, we look for the most similar reference pixel in the embedding space and assign the same label to it. The proposed method is sketched in Figure~\ref{fig:architecture}. Our method consists of two stages when processing a new video: we first embed each pixel into a $d$-dimensional embedding space using the proposed embedding network. Then the second step is to perform per-pixel retrieval in this space to transfer labels to each pixel according to its nearest reference pixel. 

A key aspect of our approach, which allows for a fast user interaction, is our way of incorporating the user input.  Alternative approaches have been exploited to inject user input into deep learning systems:

\textit{User input to fine-tune the model:} The first way is to fine-tune the network to the specific object based on the user input. For example, techniques such as OSVOS~\cite{caelles2017one} or MaskTrack~\cite{Perazzi2017} fine-tune the network at test time based on the user input. When processing a new video, they require many iterations of training to adapt the model to the specific target object. This approach can be time-consuming (seconds per sequence) and therefore impractical for real-time applications, especially with a human in the loop. 

\textit{User input as the network input:} Another way of injecting user interaction is to use it as an additional input to the network. In this way, no training is performed at test time. Such methods typically either directly concatenate the user input with the image~\cite{xu2016deep}, or use a sub-network to encode the user input~\cite{shaban2017one,yang2018efficient}. A drawback of these methods is that the network has to be recomputed once the user input changes. This can still be a considerable amount of time, especially for video, considering the large number of frames.

In contrast to previous methods, in this work user input is disentangled from the network computation, thus the forward pass of the network needs to be computed only once. The only computation after user input is then a nearest-neighbor search, which is very fast and enables rapid response to the user input.

\subsection{Segmentation as Pixel-wise Retrieval}
\label{sec:main_method}
For clarity, here we assume a single-object segmentation scenario, and the segmentation mask of first frame is used as user input.
The discussion is, however, applicable for multiple objects and for other types of inputs as well.

The task of semi-supervised video object segmentation is defined as follows: segmenting an object in a video given the object mask of the first frame. Formally, let us denote the \textit{i}-th pixel in the \textit{j}-th frame of the input video as $x_{j,i}$. The user provides the annotation for the first frame: $(x_{1,i},l_{1,i})$, where $l\in\{0,1\}$, and $l_{1,i}=0,1$ indicates $x_{1,i}$ belongs to background and foreground, respectively. We refer to these annotated pixels as reference pixels. The goal is then to infer the labels of all the unlabeled pixels in other frames $l_{j,i}$ with $j>1$.

\paragraph{Embedding Model:}\rule{0mm}{2mm}\\We build an embedding model $f$ and each pixel $x_{j,i}$ is represented as a $d$-dimensional embedding vector $e_{j,i} = f(x_{j,i})$. Ideally, pixels belonging to the same object are close to each other in the embedding space, and pixels belonging to different objects are distant to each other. In more detail, our embedding model is build on DeepLab-v2~\cite{chen2016deeplab} based on the ResNet101~\cite{He+16} backbone architecture. First, we pre-train the network for semantic segmentation on COCO~\cite{Lin2014} using the same procedure presented in~\cite{chen2016deeplab} and then we remove the final classification layer and replace it with a new convolutional layer with $d$ output channels. We fine-tune the network to learn the embedding for video object segmentation, which will be detailed in Section~\ref{sec:training}. To avoid confusion, we refer to the the original DeepLab-v2 architecture as \textit{base feature extractor} and to the two convolutional layers as \textit{embedding head}. The resulting network is fully convolutional, thus the embedding vector of all pixels in a frame can be obtained in a single forward pass. For an image of size $h\times w$ pixels the output is a tensor $[h/8,w/8,d]$, where $d$ is the dimension of the embedding space. We use $d=128$ unless otherwise specified. The tensor is $8$ times smaller due to that the network has a stride length of $8$ pixels.

Since an FCN is deployed as the embedding model, spatial and temporal information are not kept due to the translation invariance nature of the convolution operation. However, such information is obviously important for video and should not be ignored when performing segmentation. We circumvent this problem with a simple approach: we add the spatial coordinates and frame number as additional inputs to the embedding head, thus making it aware of spatial and temporal information. Formally, the embedding function can be represented as $e_{j,i} = f(x_{j,i},i,j)$, where $i$ and $j$ refer to the $i$th pixel in frame $j$. This way, spatial information $i$ and temporal information $j$ can also be encoded in the embedding vector $e_{j,i}$.

\paragraph{Retrieval with Online Adaptation:}\rule{0mm}{2mm}\\
During inference, video object segmentation is simply performed by retrieving the closer reference pixels in the embedded space. We deploy a $k$-Nearest Neighbors ($k$NN) classifier which finds the set of reference pixels whose feature vector $e_i^j$ is closer to the feature vector of the pixels to be segmented. In the experiments, we set $k=5$ for the semi-supervised case, and $k=1$ for the interactive segmentation case. Then, the identity of the pixel is computed by a majority voting of the set of closer reference pixels. Since our embedding model operates with a stride of $8$, we upsample our results to the original image resolution by the bilateral solver~\cite{barron2016fast}. 

A major challenge for semi-supervised video object segmentation is that the appearance changes as the video progresses. The appearance change causes severe difficulty for a fixed model learned in the first frame. As observed in \cite{voigtlaender17BMVC,chen2018road}, such appearance shift usually leads to a decrease in performance for FCNs. To cope with this issue, OnAVOS\cite{voigtlaender17BMVC} proposes to update the model using later frames where their prediction is very confident. In order to update their model online, however, they have to run a few iterations of the fine-tuning algortihm using highly confident samples, which makes their method even slower than the original OSVOS.

This issue can also be understood as the sample distribution shifts in the  embedding space over time. In this work, we can easily update the model online to capture the appearance change, a process that is nearly effortless. In particular we initialize the pool of reference samples with the samples that the user have annotated. As the video progresses, we gradually add samples with high confidence to the pool of reference samples. We add the samples into our reference pool if all of its $k=5$ near neighbors agree with the label.

\paragraph{Generalization to different user input modes and multiple objects:}\rule{0mm}{2mm}\\
So far we focused on single-object scenarios where user interaction is provided as the full object mask in the first frame. However, multiple object might be present in the video, and the user input might be in an arbitrary form other than the full mask of the first frame. Our method can be straightforwardly applicable to such cases.

In a general case, the input from user can be represented as a set of pixels and its corresponding label: $\{x_{i,j},l_{i,j}\}$ without need for all inputs to be on the first frame ($j=1$) or the samples to be exhaustive (covering all pixels of one frame). Please note that the latter is in contrast to the majority of semi-supervised video object segmentation techniques, which assume a full annotated frame to segment the object from the video.

In our case, the input ${x_{i,j}}$ can be in the form of clicked points, drawn scribbles, or others possibilities. The label $l_{i,j}$ can also be an integer $l_i^j\in\{1 ... K\}$ representing an identifier of an object within a set of $K$ objects, thus generalizing our algorithm to multiple-object video segmentation.

\subsection{Training}
\label{sec:training}
The basic idea of metric learning is to \textit{pull} similar samples close together and \textit{push} dissimilar points far apart in the embedding space. A proper training loss and sampling strategy are usually of critical importance to learn a robust embedding. Below we present our training loss and sampling strategy specifically designed for video object segmentation.

\begin{figure}[h]
\hspace{-4mm}
\resizebox{1.05\linewidth}{!}{%
\begin{tikzpicture}
\node (tsne) {\reflectbox{\rotatebox[origin=c]{180}{\includegraphics[height=0.15\textwidth,trim=120 220 80 220,clip]{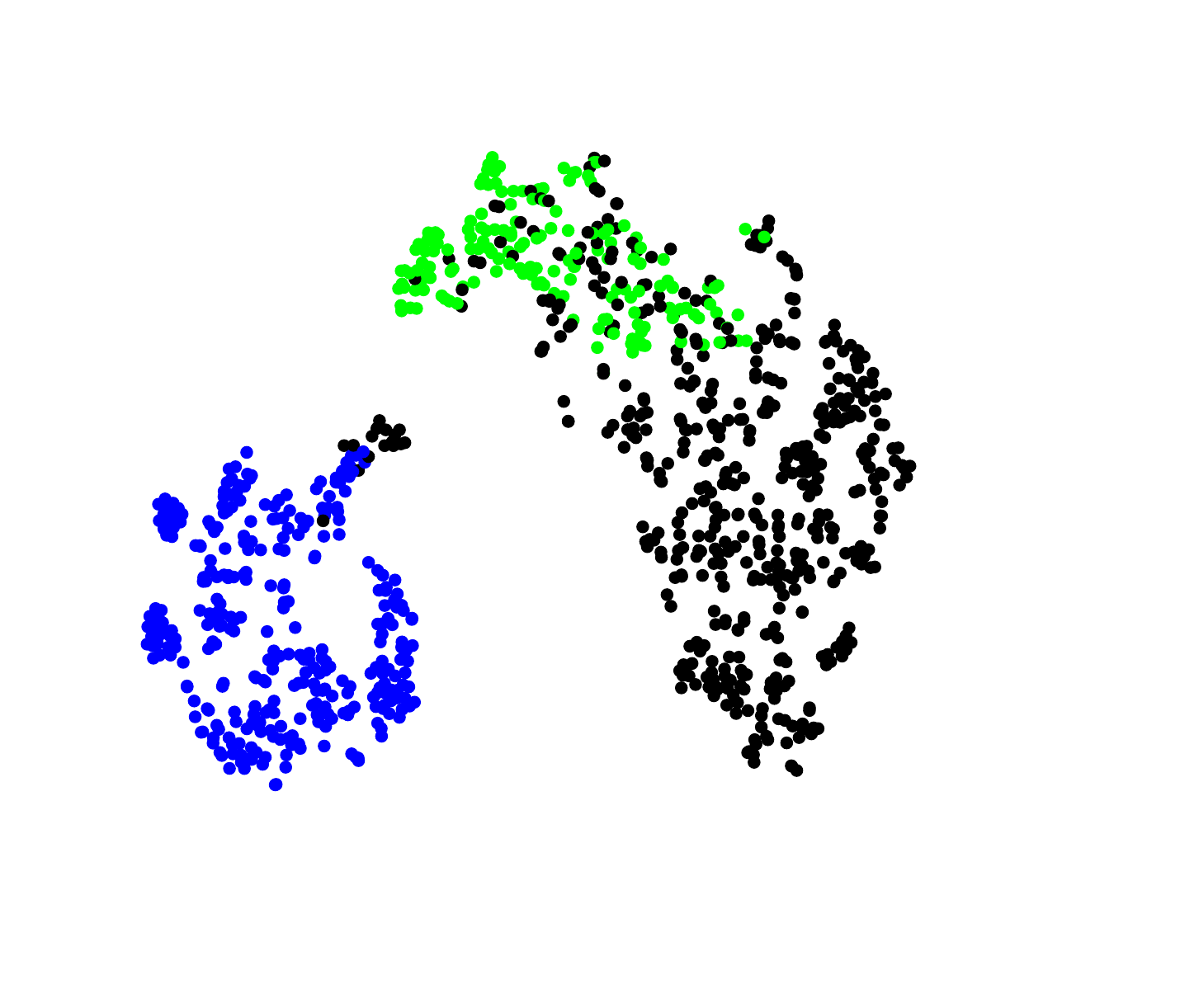}}}};
\node[anchor=west] at (tsne.east) {\includegraphics[height=0.15\textwidth]{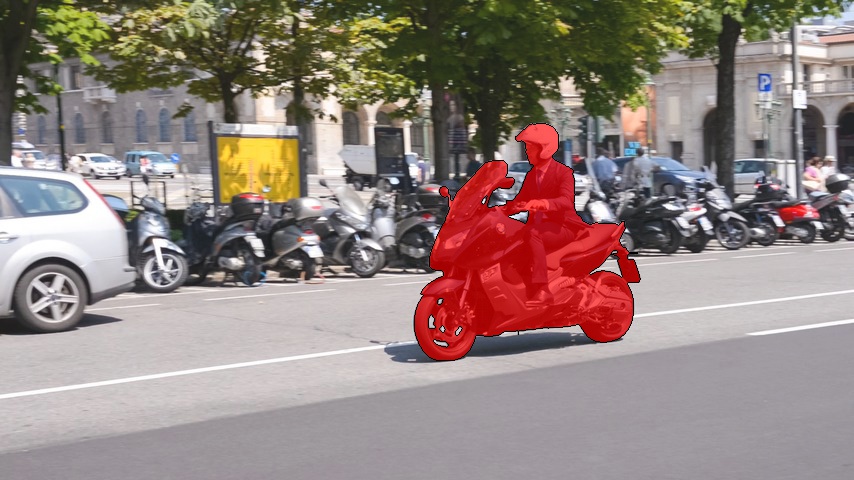}};
\draw [->, ultra thick, green] (0,-0.7) -- (4.3,-0.4);
\draw [->, ultra thick, blue] (-1,0.5) -- (4.8,0.3);
\end{tikzpicture}}
\caption{\textbf{Illustration of pixel-wise feature distribution:} Green denotes pixels from motorbike, blue represents person, and black background. The object of interest in this video and the annotation is the human and the motorbike. However, features from motorbike and person lie in two clusters in the feature space. Pulling these two cluster close might be harmful for the metric learning. Visualization is done by t-SNE~\cite{maaten2008visualizing}.}
\label{fig:features_loss}
\vspace{-5mm}

\end{figure}

\paragraph{Training loss:}\rule{0mm}{2mm}\\
In the metric learning literature, contrastive loss~\cite{chopra2005learning,hadsell2006dimensionality}, triplet loss~\cite{chechik2010large}, and their variants are widely used for metric learning. We argue, however, and verify in our experiments, that the standard losses are not suitable for the task at hand, i.e.\ video object segmentation, arguably due to the intra-object variation present in a video. In other words, the triplet loss is designed for the situation where the identity of the sample is clear, which is not the case for video object segmentation as an object can be composed of several parts, and each part might have very different appearance. Pulling these samples close to each other, therefore, is an extra constraint that can be harmful for learning a robust metric. We illustrate this effect with an example in Figure~\ref{fig:features_loss}.

\begin{table*}
\centering
\rowcolors{2}{white}{rowblue}
\resizebox{\linewidth}{!}{%
\sisetup{detect-all=true}
\begin{tabular}{llS[table-format=2.1]S[table-format=2.1]S[table-format=2.1]S[table-format=2.1]S[table-format=2.1]S[table-format=2.1]S[table-format=2.1]S[table-format=2.1]S[table-format=2.1]S[table-format=2.1]S[table-format=2.1]S[table-format=2.1]S[table-format=2.1]}
\toprule
\multicolumn{2}{c}{Measure} & \si{\nonavos} & \si{\nosvos} & \si{\nmsk} & \si{\npml} & \si{\nsfls} & \si{\nctn} & \si{\nvpn} & \si{\nofl} & \si{\nbvs} & \si{\nfcp} & \si{\njmp} & \si{\nhvs} & \si{\nsea} \\
\cmidrule(lr){1-2} \cmidrule(lr){3-15} 
\cellcolor{rowblue}$\mathcal{J}\&\mathcal{F}$  & Mean $\mathcal{M} \uparrow$   &\bfseries 85.5 &          80.2 &          77.5 &          77.4 &          76.0 &          71.4 &          67.8 &          65.7 &          59.4 &          53.8 &          55.1 &          53.8 &          49.2 \\
\hline
                                 & Mean $\mathcal{M} \uparrow$     &\bfseries 86.1 &          79.8 &          79.7 &          75.5 &          76.1 &          73.5 &          70.2 &          68.0 &          60.0 &          58.4 &          57.0 &          54.6 &          50.4 \\
\cellcolor{white}$\mathcal{J}$ & Recall $\mathcal{O} \uparrow$   &\bfseries 96.1 &          93.6 &          93.1 &          89.6 &          90.6 &          87.4 &          82.3 &          75.6 &          66.9 &          71.5 &          62.6 &          61.4 &          53.1 \\
                                 & Decay $\mathcal{D} \downarrow$  &          5.2 &          14.9 &          8.9 &          8.5 &          12.1 &          15.6 &          12.4 &          26.4 &          28.9 &\bfseries -2.0 &          39.4 &          23.6 &          36.4 \\
\hline
                                 & Mean $\mathcal{M} \uparrow$     &\bfseries 84.9 &          80.6 &          75.4 &          79.3 &          76.0 &          69.3 &          65.5 &          63.4 &          58.8 &          49.2 &          53.1 &          52.9 &          48.0 \\
\cellcolor{rowblue}$\mathcal{F}$ & Recall $\mathcal{O} \uparrow$   &          89.7 &          92.6 &          87.1 &\bfseries 93.4 &          85.5 &          79.6 &          69.0 &          70.4 &          67.9 &          49.5 &          54.2 &          61.0 &          46.3 \\
                                 & Decay $\mathcal{D} \downarrow$  &          5.8 &          15.0 &          9.0 &          7.8 &          10.4 &          12.9 &          14.4 &          27.2 &          21.3 &\bfseries -1.1 &          38.4 &          22.7 &          34.5 \\
\hline
\cellcolor{white}$\mathcal{T}$ & Mean $\mathcal{M} \downarrow$   &          19.0 &          37.8 &          21.8 &          47.0 &          18.9 &          22.0 &          32.4 &          22.2 &          34.7 &          30.6 &          15.9 &          36.0 &\bfseries 15.4 \\
\bottomrule
\end{tabular}
}
\vspace{2mm}
\caption{\textbf{Evaluation results on DAVIS 2016 validation set set:} We compare the proposed method with an exhaustive set of very recent techniques. }
\label{tab:global_table}
\end{table*}

Keeping this in mind, we modify the standard triplet loss to adapt it to our application. Formally, let us refer to anchor sample as $x^a$. $x^p\in \mathcal{P}$ is a positive sample from a positive sample pool $\mathcal{P}$. Similarly, $x^n$ denotes a negative sample and $\mathcal{N}$ denotes the negative pool. The standard triplet loss pushes the negative points further away than the distance between anchor and positive points. Since we do not want to pull every pair of positive points close (different parts of an object that look different), we modify the loss to only push the smallest negative points further than the smallest positive points, the loss can thus be represented as:

\begin{equation}
\sum_{x^a\in  \mathcal{A}} \{\min_{x^p\in \mathcal{P}}\| f(x^a) - f(x^p) \|^2_2 - \min_{x^n\in \mathcal{N}}\| f(x^a) - f(x^n) \|^2_2 +\alpha\}
\label{eqn:loss_pixel_retrieval}
\end{equation}
where $\alpha$ is the slack variable to control the margin between positive and negative samples, as in the standard formulation, and we denote the set of anchors as $\mathcal{A}$. 

For each anchor sample $x^a$ we have two pools of samples: one pool of positive samples $\mathcal{P}$, whose labels are consistent with the anchor and another pool of negative examples $\mathcal{N}$, whose labels are different from the anchor sample. We take the closest sample to the anchor in each pool, and we compare the positive distance and negative distance. Intuitively, the loss \textit{pushes} only the closest negative away, while keeping the closest positive closer. 

\paragraph{Training Strategy:}\rule{0mm}{2mm}\\
During training, we have fully annotated videos available (object segmentation on each frame). To form a valid triplet to train from, to be used in the aforementioned loss, we need to sample an anchor point $x^a$, a positive sample pool $\mathcal{P}$ and a negative sample pool $\mathcal{N}$. For this purpose, three frames are randomly sampled from the training video: from one we sample anchor points and the pixels from the other two frames are joined together. From those, the pixels that have the same label than the anchor form the positive pool $\mathcal{P}$, and the rest form the negative pool $\mathcal{N}$. Note that the pools are sampled from two different frames to have temporal variety, which is needed for the embedding head to learn to weight the temporal information from the feature vector. Also, we do not use pixels from the the anchor frame in the pools to avoid too easy samples.

In each iteration, a forward pass is performed on three randomly selected frames with one frame as the anchor. Then the anchor frame is used to sample 256 anchor samples, and the positive and negative pools are all foreground and background pixels in the other two frames. We compute the loss according to Equation~\ref{eqn:loss_pixel_retrieval} and the network is trained in an end to end manner.

\section{Experimental Validation}
We evaluate the proposed method mainly on DAVIS 2016~\cite{Perazzi2016}, a dataset containing 50 full high-definition videos annotated with pixel-level accurate object masks (one per sequence) densely on all the frames. We train our model on the 30 training videos and report the results on the validation set, consisting of 20 videos. We perform experiments with multiple objects in DAVIS 2017~\cite{Pont-Tuset_arXiv_2017}, an extension of the former to 150 sequences and multiple objects.

\subsection{Semi-supervised VOS on DAVIS}
We first consider the semi-supervised scenario defined in DAVIS 2016, where the methods are given the full segmentation of the object in the first frame and the goal is to segment the rest of the frames.

\begin{figure*}
\pgfplotstableread{data/per_seq_mean_JF.txt}\perseqdata
\mbox{%
\begin{minipage}{0.88\textwidth}
  \resizebox{\textwidth}{!}{%
    \begin{tikzpicture}
        \begin{axis}[set layers,width=1.4\textwidth,height=0.4\textwidth,
                ybar=0pt,bar width=0.10,
                grid=both,
				grid style=dotted,
                minor ytick={0,0.05,...,1.1},
    			ytick={0,0.1,...,1.1},
			    yticklabels={0,.1,.2,.3,.4,.5,.6,.7,.8,.9,1},
                ymin=0, ymax=1,
                xtick = data, x tick label style={rotate=20,anchor=north east,xshift=7pt,yshift=5pt},
                xticklabels from table={\perseqdata}{Seq},
                major x tick style = transparent,
                enlarge x limits=0.03,
                font=\scriptsize,
                ]
             \addplot[draw opacity=0,fill=Set2-8-1,mark=none,legend image post style={yshift=-0.1em}] table[x expr=\coordindex,y=onavos]{\perseqdata};
            \label{fig:perseq:onavos}
            \addplot[draw opacity=0,fill=Set2-8-2,mark=none,legend image post style={yshift=-0.1em}] table[x expr=\coordindex,y=osvos]{\perseqdata};
            \label{fig:perseq:osvos}
            \addplot[draw opacity=0,fill=Set2-8-3,mark=none,legend image post style={yshift=-0.1em}] table[x expr=\coordindex,y=msk]{\perseqdata};
            \label{fig:perseq:msk}    
			\addplot[draw opacity=0,fill=Set2-8-8,mark=none,legend image post style={yshift=-0.1em}] table[x expr=\coordindex,y=sfls]{\perseqdata};
			\label{fig:perseq:sfls}
            \addplot[draw opacity=0,fill=Set2-8-4,mark=none,legend image post style={yshift=-0.1em}] table[x expr=\coordindex,y=ctn]{\perseqdata};
            \label{fig:perseq:ctn}
            \addplot[draw opacity=0,fill=Set2-8-5,mark=none,legend image post style={yshift=-0.1em}] table[x expr=\coordindex,y=vpn]{\perseqdata};
            \label{fig:perseq:vpn}
            \addplot[draw opacity=0,fill=Set2-8-6,mark=none,legend image post style={yshift=-0.1em}] table[x expr=\coordindex,y=ofl]{\perseqdata};
            \label{fig:perseq:ofl}
            \addplot[draw opacity=0,fill=Set2-8-7,mark=none,legend image post style={yshift=-0.1em}] table[x expr=\coordindex,y=bvs]{\perseqdata};
			\label{fig:perseq:bvs}

            \addplot[blue,sharp plot,update limits=false,mark=*,mark size=1,line width=1.2pt, legend image post style={yshift=-0.4em}] table[x expr=\coordindex,y=pml]{\perseqdata};
            \label{fig:perseq:pml}
            
            \addplot[black,sharp plot,update limits=false] coordinates{(-0.5,0) (20.5,0)};
        \end{axis}
    \end{tikzpicture}
    }
    \end{minipage}
    \hspace{0mm}
    \begin{minipage}{0.07\textwidth}
    \scriptsize
    \begin{tabular}{@{}l@{\hspace{1.5mm}}l}
    \ref{fig:perseq:pml}& PML (Ours)\\
    \ref{fig:perseq:onavos} &\nonavos~\cite{\onavos}\\
    \ref{fig:perseq:osvos} &\nosvos{}~\cite{\osvos}\\
    \ref{fig:perseq:msk} &\nmsk~\cite{\msk}\\
    \ref{fig:perseq:sfls} &\nsfls~\cite{\sfls}\\
    \ref{fig:perseq:ctn} &\nctn~\cite{\ctn}\\
    \ref{fig:perseq:vpn} &\nvpn~\cite{\vpn}\\
    \ref{fig:perseq:ofl} &\nofl~\cite{\ofl}\\
    \ref{fig:perseq:bvs} &\nbvs~\cite{\bvs}
   \end{tabular}
    \end{minipage}
    }
    \vspace{-1mm}
    \caption{\label{fig:perseq}: Per-sequence results of mean region similarity ($\J$) and contour accuracy ($\F$).
    The rest of the state-of-the-art techniques are shown using bars, ours is shown using a line. Sequences are sorted by our performance.}
\end{figure*}
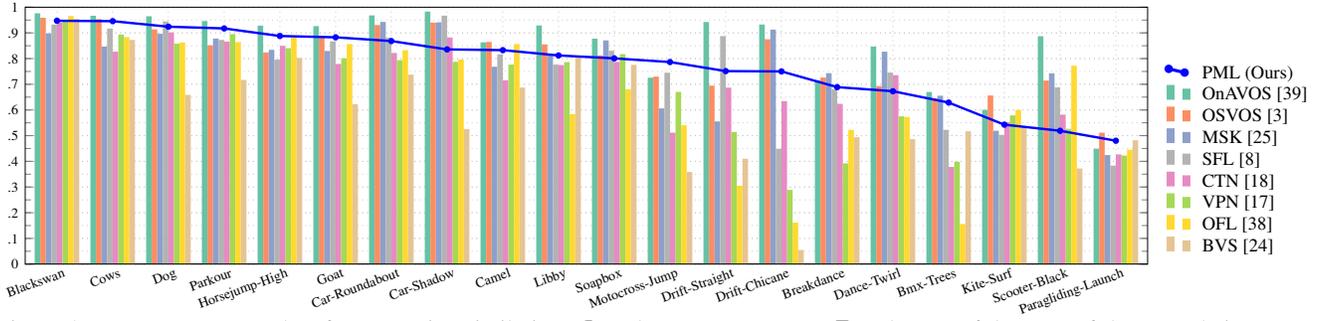

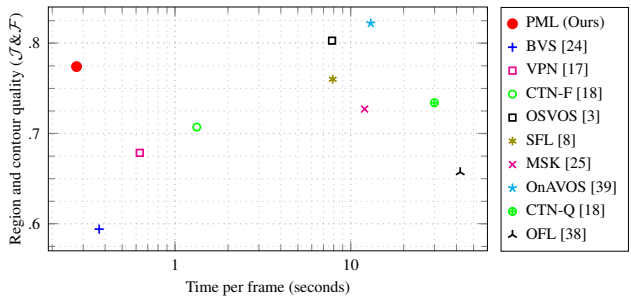
\begin{figure}
\centering
\resizebox{\linewidth}{!}{\begin{tikzpicture}[/pgfplots/width=1\linewidth, /pgfplots/height=0.64\linewidth, /pgfplots/legend pos=south east]
    \begin{axis}[ymin=0.57,ymax=0.84,xmin=0.19,xmax=60,enlargelimits=false,
        xlabel=Time per frame (seconds),
        ylabel=Region and contour quality ($\J$\&$\F$),
		font=\scriptsize,
        grid=both,
		grid style=dotted,
        xlabel shift={-2pt},
        ylabel shift={-5pt},
        xmode=log,
        legend columns=1,
        minor ytick={0,0.025,...,1.1},
        ytick={0,0.1,...,1.1},
		yticklabels={0,.1,.2,.3,.4,.5,.6,.7,.8,.9,1},
	    xticklabels={.1,1,10,100},
        legend pos= outer north east,
        legend cell align={left}
        ]

		\addplot[red,mark=*,only marks,line width=0.75] coordinates{(0.275,0.774)};
        \addlegendentry{\hphantom{i}\npml}
        \label{fig:qual_vs_time:pml}
        
        \addplot[blue,mark=+,only marks,line width=0.75] coordinates{(0.37,0.594)};
        \addlegendentry{\hphantom{i}\nbvs~\cite{\bvs}}
        \label{fig:qual_vs_time:bvs}
       
        \addplot[magenta,mark=square,only marks,line width=0.75, mark size=1.45] coordinates{(0.63,0.6785)};
        \addlegendentry{\hphantom{i}\nvpn~\cite{\vpn}}
        
        \addplot[green,mark=o, mark size=1.6,only marks, line width=0.75] coordinates{(1.33,0.707)};
        \addlegendentry{\hphantom{i}\nctn-F~\cite{\ctn}}
      
              \addplot[black,mark=square,only marks,line width=0.75, mark size=1.45] coordinates{(7.832060,	0.802710)};
        \addlegendentry{\hphantom{i}\nosvos~\cite{\osvos}}
        
        \addplot[olive,mark=asterisk, mark size=1.9,only marks, line width=0.75] coordinates{(7.9,0.76)};
        \addlegendentry{\hphantom{i}\nsfls~\cite{\sfls}}
        
        \addplot[magenta,mark=x, mark size=2.1,only marks, line width=0.75]coordinates{(12,0.727)};
        \addlegendentry{\hphantom{i}\nmsk~\cite{\msk}}
                
        \addplot[cyan,mark=star, mark size=2,only marks, line width=0.75] coordinates{(13,0.822)};
        \addlegendentry{\hphantom{i}\nonavos~\cite{\onavos}}

        \addplot[green,mark=oplus, mark size=1.6,only marks, line width=0.75] coordinates{(29.95,0.734)};
        \addlegendentry{\hphantom{i}\nctn-Q~\cite{\ctn}}
                
        \addplot[black,mark=Mercedes star, mark size=2.2,only marks, line width=0.75] coordinates{(42,0.657)};
        \addlegendentry{\hphantom{i}\nofl~\cite{\ofl}}
		\label{fig:qual_vs_time:ofl}

    \end{axis}
\end{tikzpicture}}
\vspace{-5mm}
   \caption{\textbf{Quality versus timing in DAVIS 2016}: $\J$\&$\F$ of all techniques with respect to their time to process one frame.
   The timing is taken from each paper. \nonavos{} and \nmsk{} do not report their timings with the post-processing steps that lead to the most accurate results, so we compare to the version with reported times.}
   \label{fig:qual_vs_time}
   \vspace{-3mm}
\end{figure}

We compare against an exhaustive set of very recent techniques: \nonavos~\cite{\onavos}, \nosvos~\cite{\osvos}, \nmsk~\cite{\msk}, \nsfls~\cite{\sfls}, \nctn~\cite{\ctn}, \nvpn~\cite{\vpn}, \nofl~\cite{\ofl}, \nbvs~\cite{\bvs}, \nfcp~\cite{\fcp}, \njmp~\cite{\jmp}, \nhvs~\cite{\hvs}, and \nsea~\cite{\sea}; using the pre-computed results available on the DAVIS website and the metrics proposed in DAVIS ($\J$ Jaccard index or IoU, $\F$ boundary accuracy, $\T$ temporal stability). Readers are referred to each paper for more details. 

Table~\ref{tab:global_table} shows the comparison to the rest of the state of the art, i.e.\ at the best-performing regime (and slowest) of all techniques. In global terms ($\J$\&$\F$), \npml{} is comparable to \nmsk{} and only behind \nosvos{} and \nonavos{}, which are significantly slower, as we will show in the next experiment. Our technique is especially competitive in terms of boundary accuracy ($\F$), despite there is no refinement or smoothing step explicitly tackling this feature as in other methods.

To analyze the trade off between quality and performance, Figure~\ref{fig:qual_vs_time} plots the quality of each technique with respect to their mean time to process one frame (in 480p resolution). Our technique presents a significantly better trade off than the rest of techniques. Compared to the fastest one (\nbvs), we perform +18 points better while still being 100 milliseconds faster. Compared to the technique with more accurate results (\nonavos{}), we lose 5 points but we process each frame 43$\times$ faster.

Figure~\ref{fig:perseq} breaks the performance into each of the 20 sequences of DAVIS validation. We can observe that we are close to the best performance in the majority of the sequences, we obtain the best result in some of them, and our worst performance is 0.5, which shows the robustness of the embedding over various challenges and scenarios.

\begin{figure*}
\centering
\resizebox{0.98\textwidth}{!}{%
	  \setlength{\fboxsep}{0pt}
      \rotatebox{90}{\phantom{pAAA}Cows}
      \fbox{\includegraphics[width=0.3\textwidth]{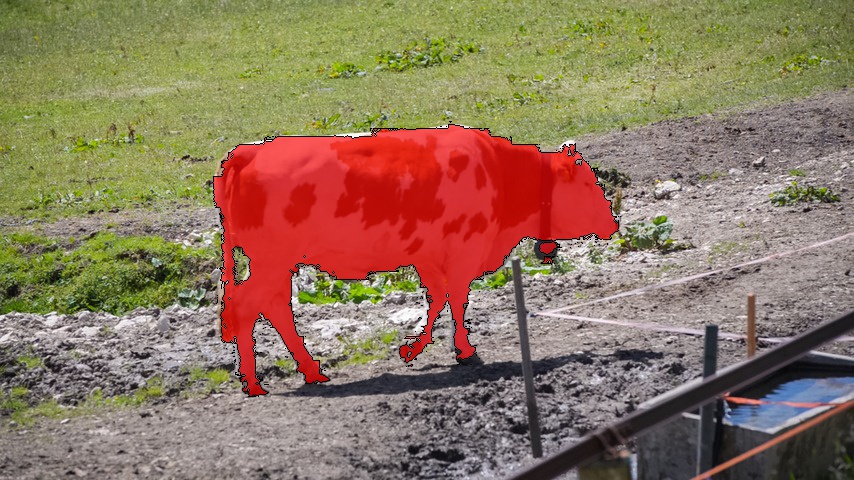}}   
      \fbox{\includegraphics[width=0.3\textwidth]{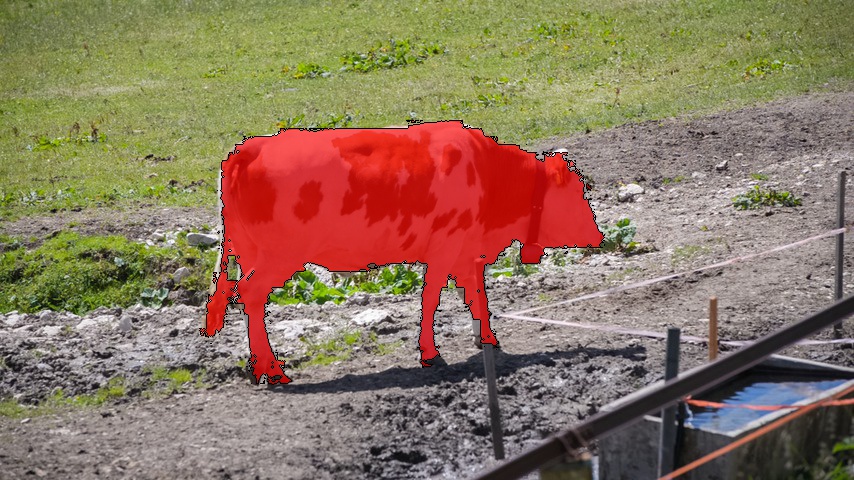}}
      \fbox{\includegraphics[width=0.3\textwidth]{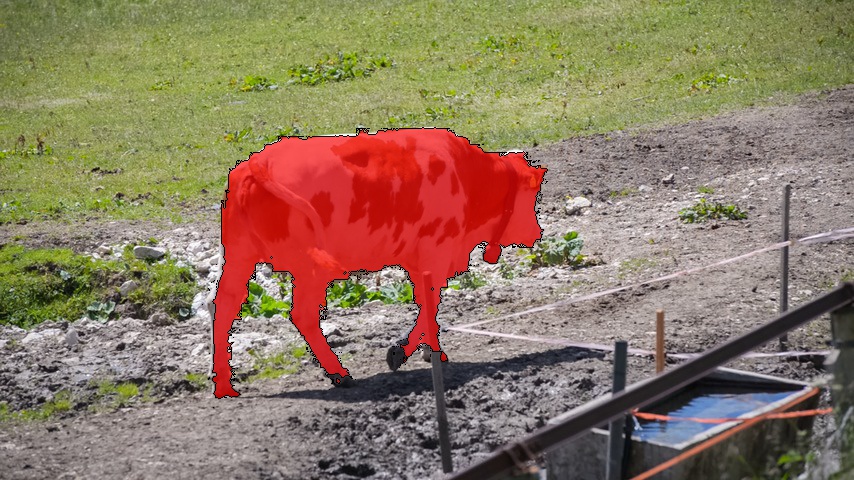}}
      \fbox{\includegraphics[width=0.3\textwidth]{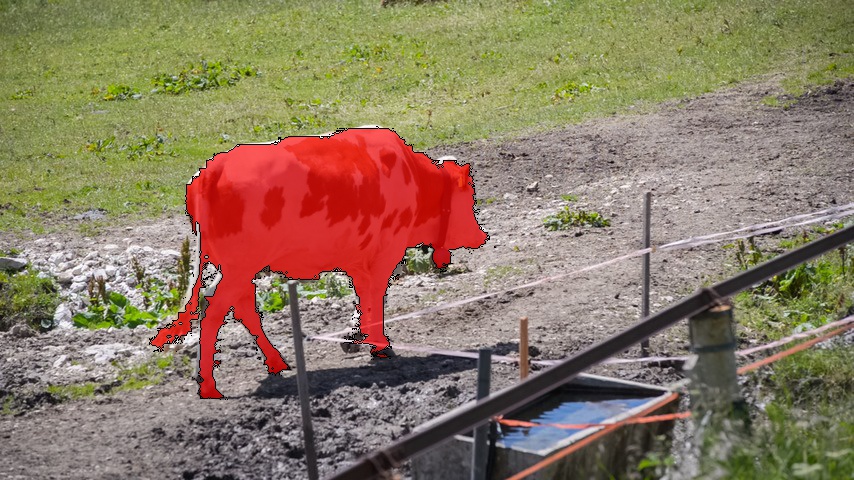}}
      \fbox{\includegraphics[width=0.3\textwidth]{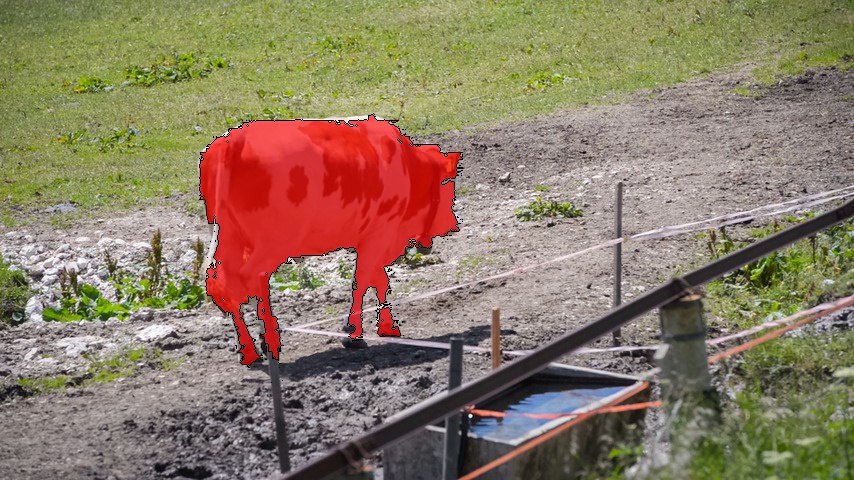}}
      }\\[1mm]
\resizebox{0.98\textwidth}{!}{%
	  \setlength{\fboxsep}{0pt}
      \rotatebox{90}{\phantom{pAAA}Parkour}
      \fbox{\includegraphics[width=0.3\textwidth]{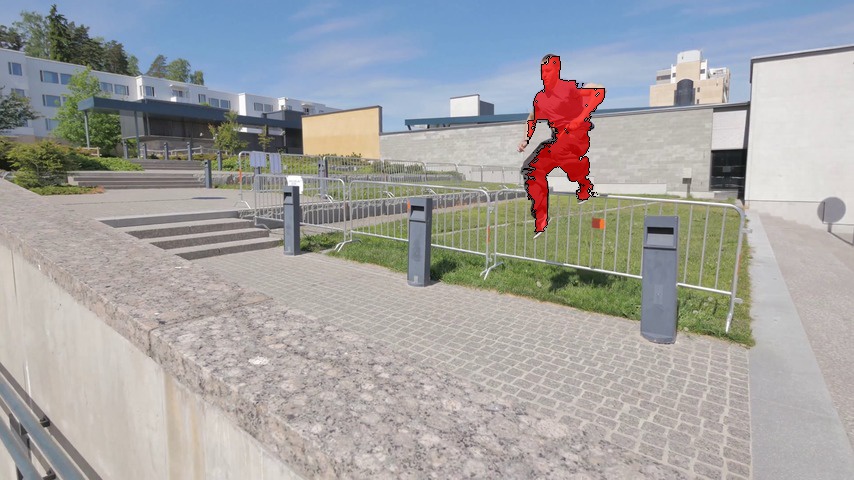}}   
      \fbox{\includegraphics[width=0.3\textwidth]{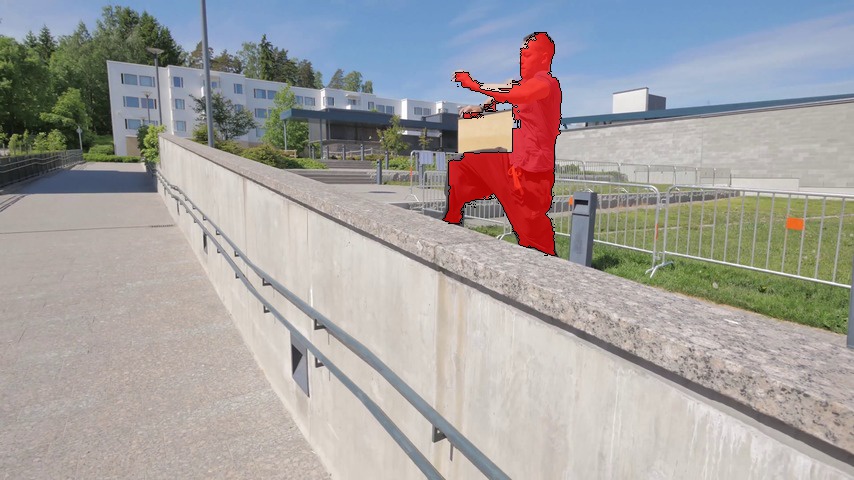}}
      \fbox{\includegraphics[width=0.3\textwidth]{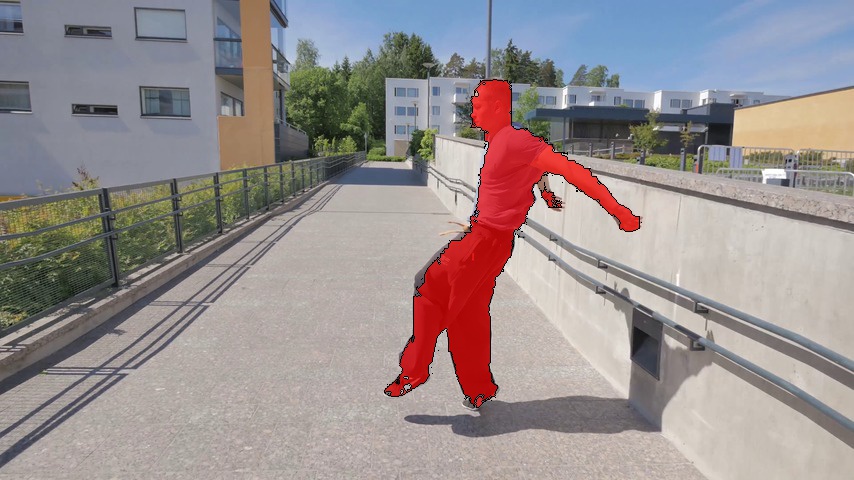}}
      \fbox{\includegraphics[width=0.3\textwidth]{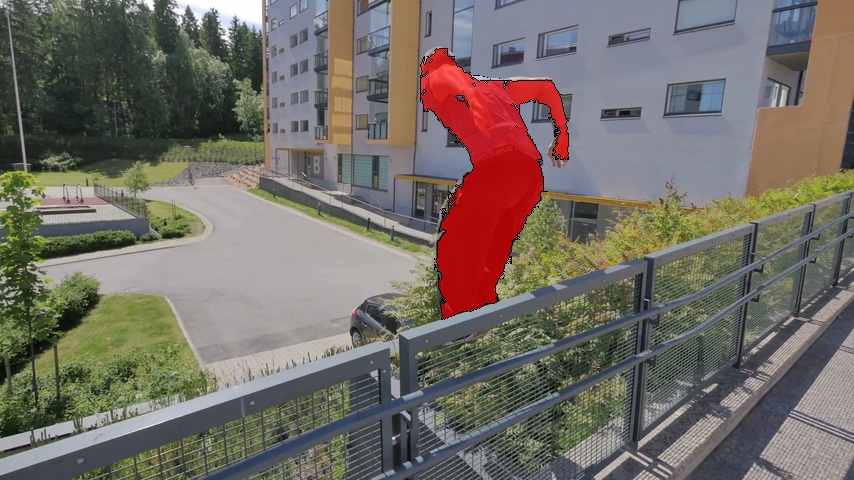}}
      \fbox{\includegraphics[width=0.3\textwidth]{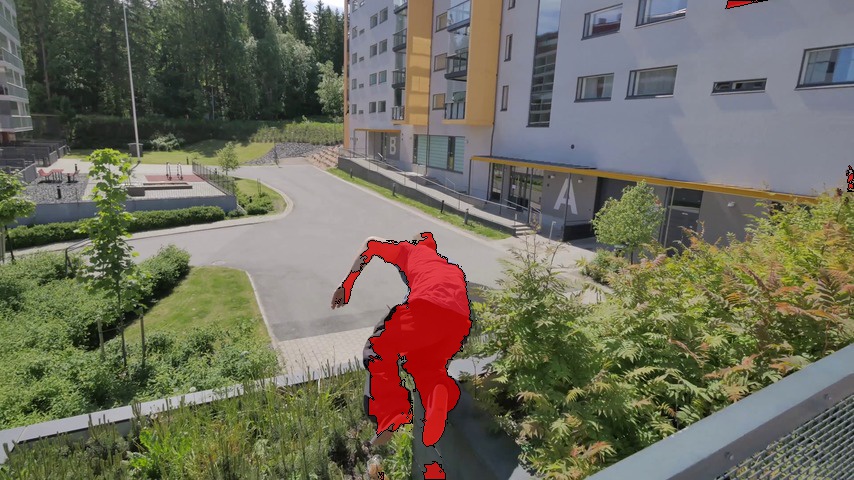}}
      }\\[1mm]
            \resizebox{0.98\textwidth}{!}{%
	  \setlength{\fboxsep}{0pt}
      \rotatebox{90}{\phantom{pAAA}Libby}
      \fbox{\includegraphics[width=0.3\textwidth]{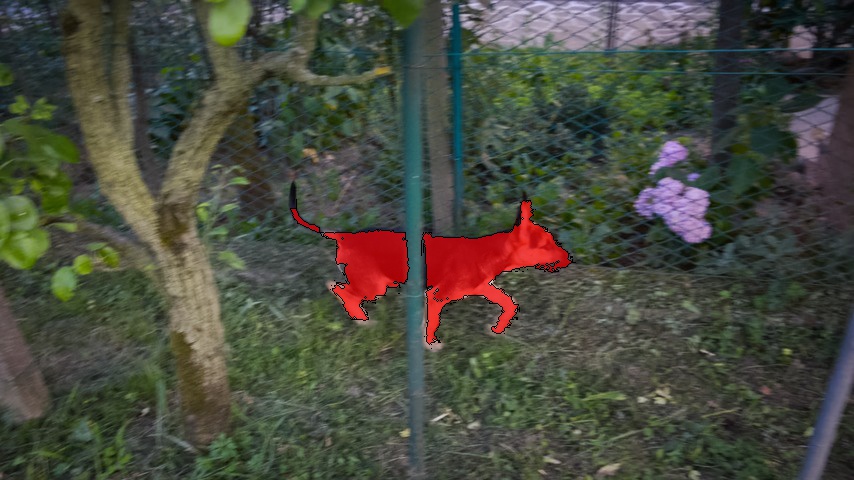}}   
      \fbox{\includegraphics[width=0.3\textwidth]{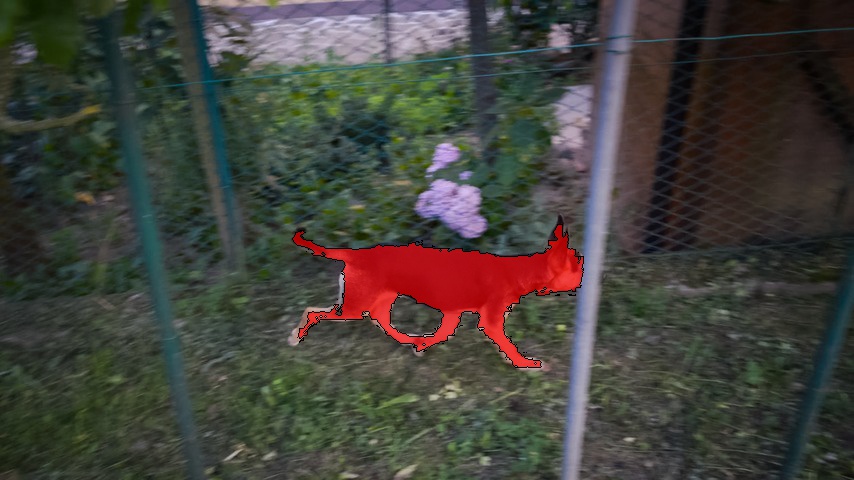}}
      \fbox{\includegraphics[width=0.3\textwidth]{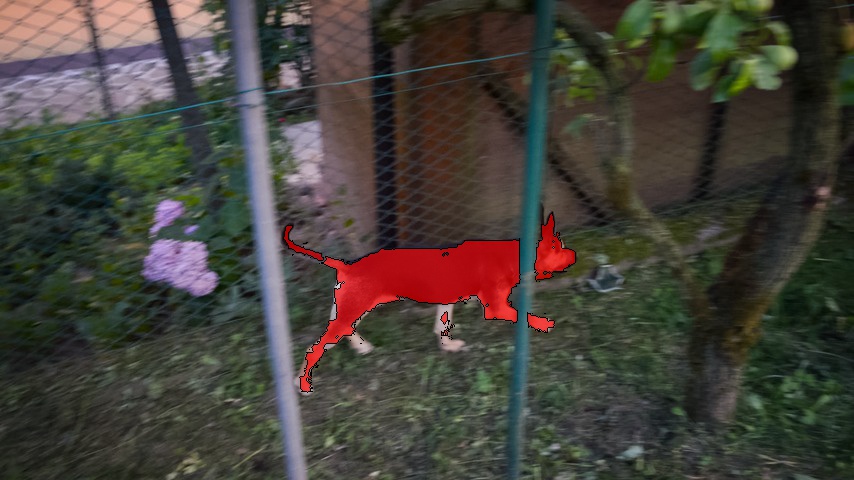}}
      \fbox{\includegraphics[width=0.3\textwidth]{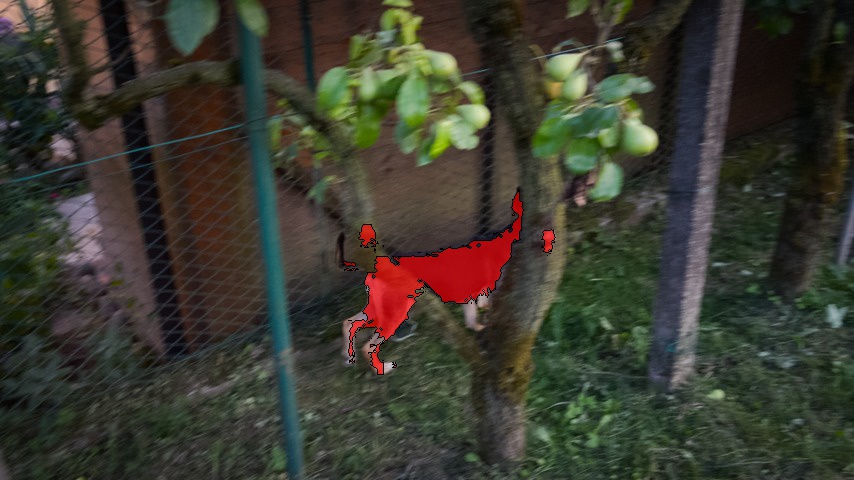}}
      \fbox{\includegraphics[width=0.3\textwidth]{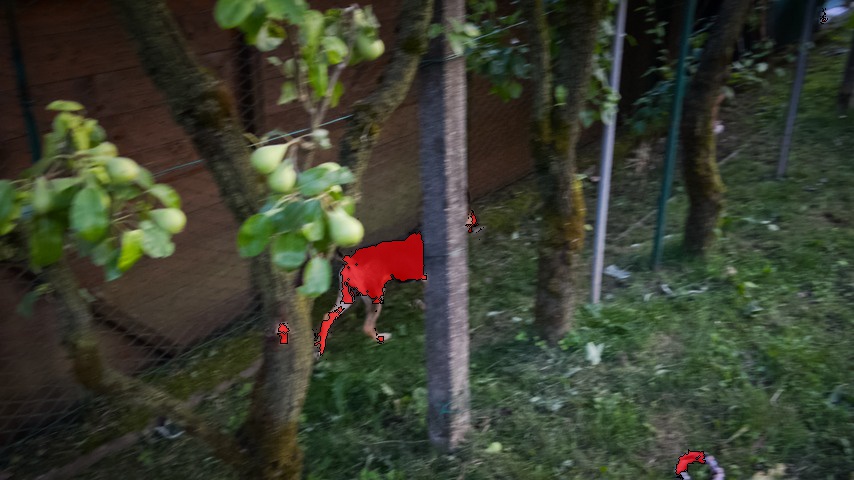}}
      }\\[1mm]
      \resizebox{0.98\textwidth}{!}{%
	  \setlength{\fboxsep}{0pt}
      \rotatebox{90}{\hspace{3mm}Motocross-Jump}
      \fbox{\includegraphics[width=0.3\textwidth]{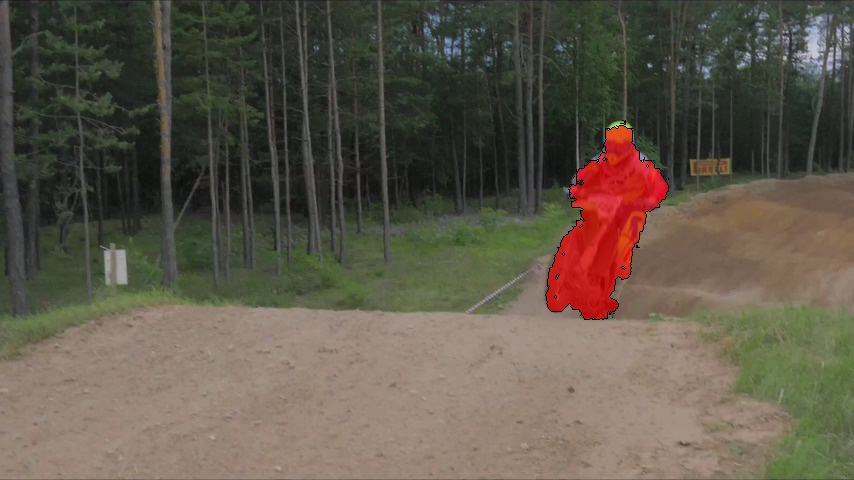}}   
      \fbox{\includegraphics[width=0.3\textwidth]{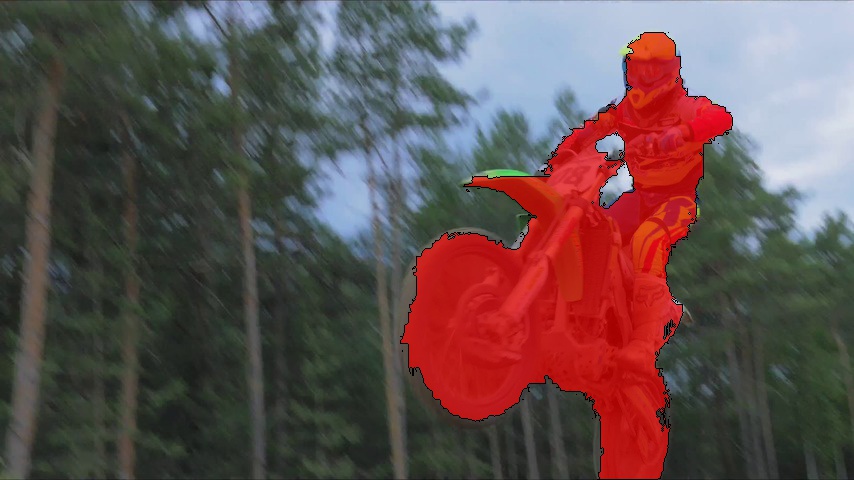}}
      \fbox{\includegraphics[width=0.3\textwidth]{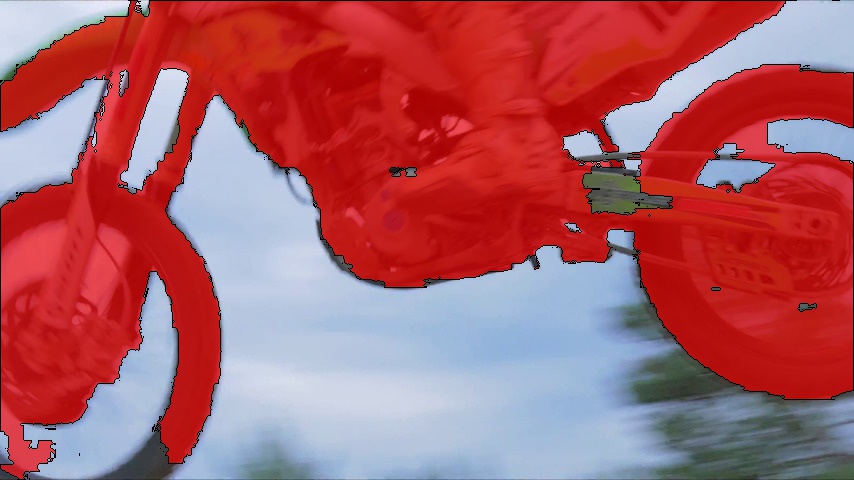}}
      \fbox{\includegraphics[width=0.3\textwidth]{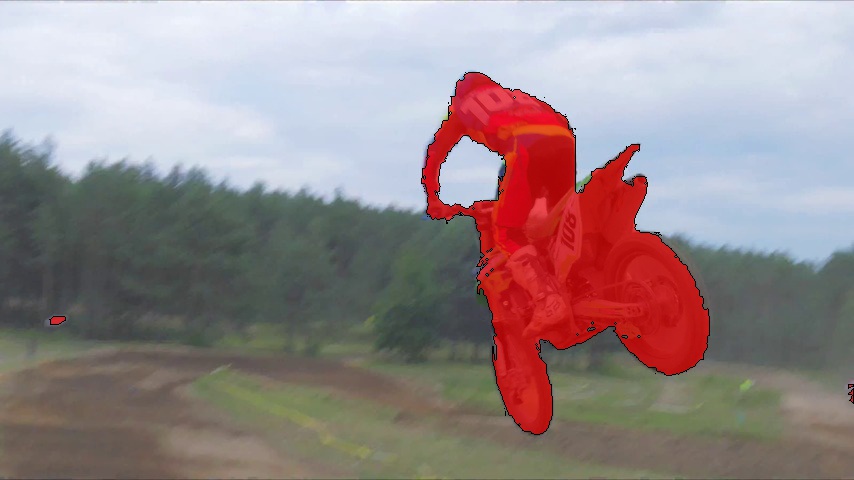}}
      \fbox{\includegraphics[width=0.3\textwidth]{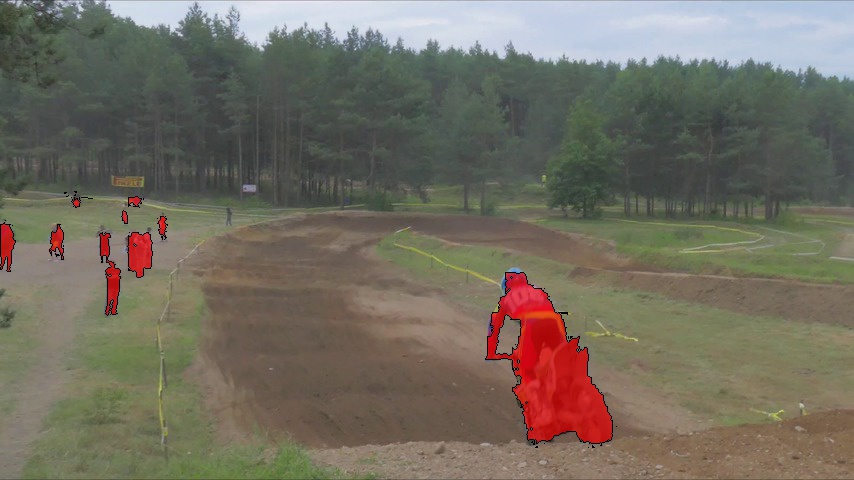}}
      }\\[1mm]
\resizebox{0.98\textwidth}{!}{%
	  \setlength{\fboxsep}{0pt}
      \rotatebox{90}{\phantom{pAA}Bmx-Trees}
      \fbox{\includegraphics[width=0.3\textwidth]{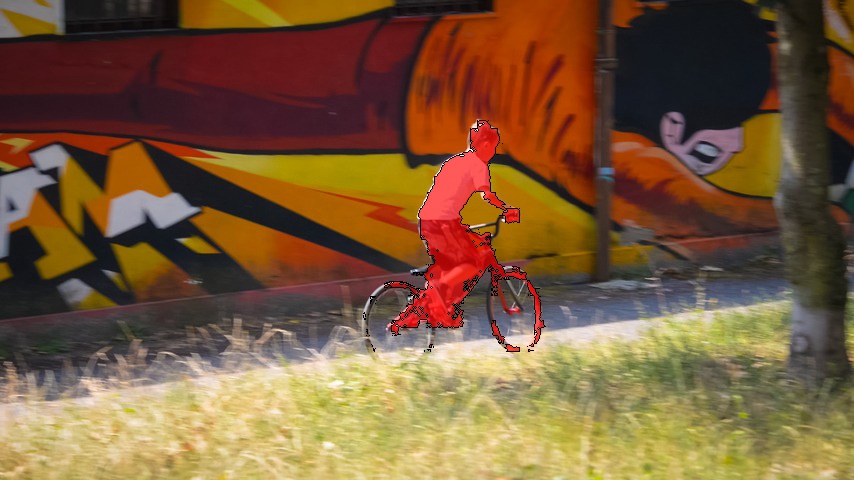}}
      \fbox{\includegraphics[width=0.3\textwidth]{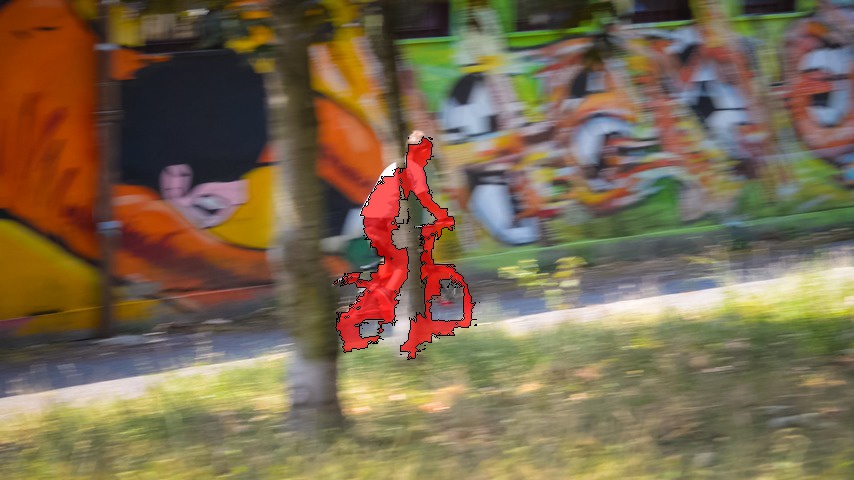}}
      \fbox{\includegraphics[width=0.3\textwidth]{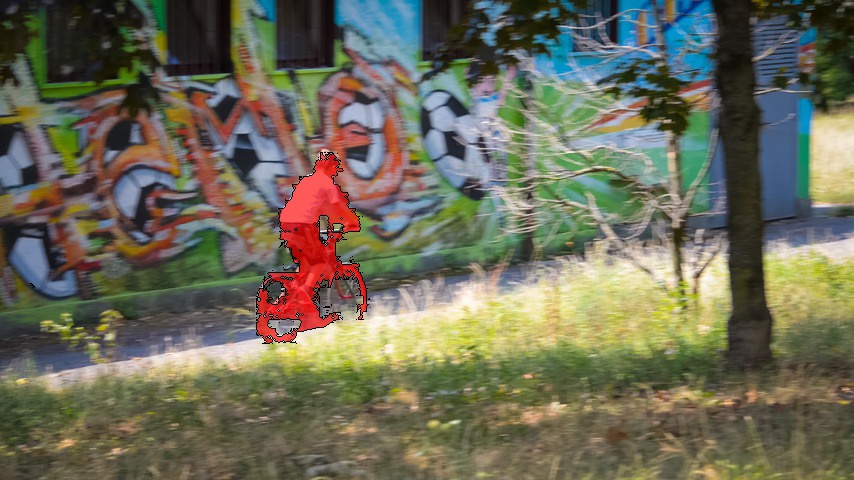}}
      \fbox{\includegraphics[width=0.3\textwidth]{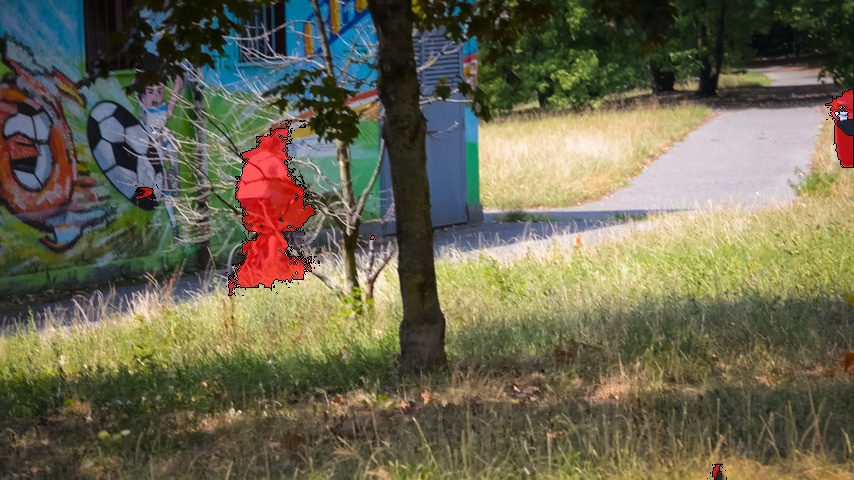}}
      \fbox{\includegraphics[width=0.3\textwidth]{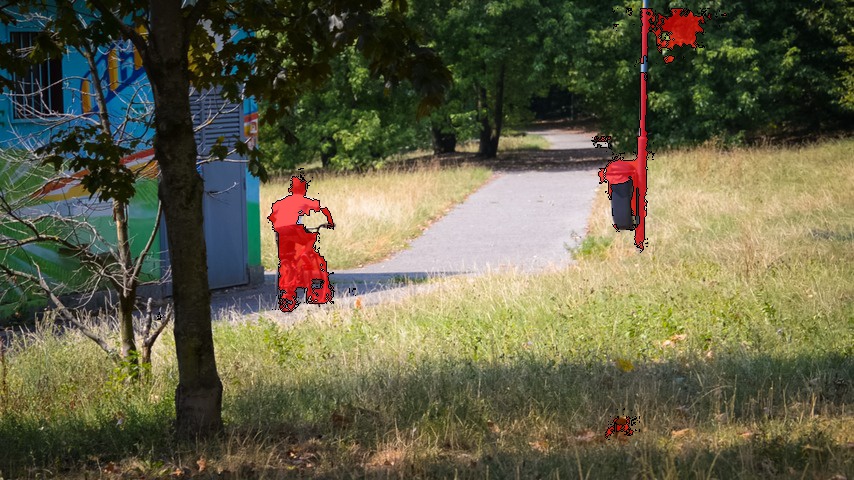}}
      }
\caption{\textbf{Qualitative results}: Homogeneous sample of DAVIS sequences with our result overlaid.}
\label{fig:qualitative}
\vspace{-2mm}
\end{figure*}

Figure~\ref{fig:qualitative} displays the qualitative results of our technique on a homogeneous set of sequences, from the ones in which we perform the best to those more challenging. Please note that in sequences Bmx-Trees (last row) and Libby (third row), our method is very robust to heavy occlusions, which is logical since we do not perform any type of temporally-neighboring propagation. Results also show that our method is robust to drastic changes in foreground scale and appearance (Motocross-Jump - fourth row) and to background appearance changes (Parkour - second row). Sequences Motocross-Jump, and BMX-Trees (fourth, and last row) show a typical failure mode (which is also observed in other techniques such as OSVOS) in which foreground objects that were not seen in the first frames are classified as foreground when they appear.

\subsection{Ablation Study}
In this section we analyze the relative importance of each proposed component, by evaluating ablated versions of our method. 

\begin{table}
\resizebox{\linewidth}{!}{%
\begin{tabular}{ cccc}
\toprule
 & Contrastive Loss & Triplet Loss & Proposed Loss \\ \midrule
Mean $\J$ & 66.1 & 69.5 & \bf75.5 \\ 
Mean $\F$ & 68.5 & 73.5 & \bf79.3 \\ \midrule
Mean $\J\&\F$& 67.3 & 71.5 & \bf77.4 \\\bottomrule
\end{tabular}}
\vspace{1pt}
\caption{Ablation Study on Different Losses}
\label{tab:ab_loss}
\end{table}

\paragraph{Training Losses for Metric Learning:}\rule{0mm}{2mm}\\
As discussed in Section~\ref{sec:training}, our embedding model is optimized using a modified version of the triplet loss. To verify the design, we compare our model with two others trained with the original contrastive loss and triplet loss, respectively; while keeping the other settings unchanged.
First, we briefly describe the different losses tested:

The \textbf{contrastive loss} operates on pairs of samples and can be written as:
\begin{equation}
L_{contra} = \sum_i^N \left\{ (y) d^2 + (1-y)\max(\alpha\!-\!d,0)^2\right\}
\nonumber
\end{equation}
where $y$ is the label of the pair ($y=0$ indicates that the pairs have different identities and $y=1$ otherwise), $d = \|x_i - x_j\|$ is the distance between two points, and $\alpha$ is a slack variable to avoid negative points being overly penalized. The loss minimizes the distance between samples if $y=1$, and maximizes it if $y=0$.

The \textbf{triplet loss} shares a similar spirit with contrastive loss, but using three samples as a unit.
Each triplet is composed of three samples: one as anchor $x_a$, one positive $x_p$, and one negative $x_n$. The positive (negative) sample has the same (different) label than the anchor.
The loss is then defined as:
\begin{equation}
L = \sum_i^N \left\{\| f(x_a) - f(x_p) \|^2_2 - \| f(x_a) - f(x_n) \|^2_2 +\alpha\right\}
\nonumber
\end{equation}
where again $\alpha$ is a slack variable to control the margin.

Table~\ref{tab:ab_loss} compares our embedding model with the models trained with the alternative losses. The results clearly show that the proposed loss achieves better performance than the alternatives.

\paragraph{Spatio-Temporal-Aware Embedding and Online Adaptation:}\rule{0mm}{2mm}\\
We proceed with our ablation analysis by studying the individual impact of two major sub-components: online adaptation and spatial and temporal awareness, as presented in Section~\ref{sec:main_method}. 

Table~\ref{tab:ab_spatial_online_adapt} presents our ablation study on each component: online adaptation provides a slight boost of $+1.2\%$ in $\J$. Bringing in spatial and temporal information gives $+2.3\%$ improvement in $\J$ and $+4.5\%$ in $\F$ which validates the importance of spatial and temporal information for video object segmentation. Combining both results gives the best performance of $75.5\%$ in overlap, which is overall $+3.5\%$ higher at nearly no extra cost. 

\begin{table}
\centering
\resizebox{\linewidth}{!}{%
\begin{tabular}{ c c c c c}
\toprule
 Spat.-Temp. & Online Adapt. & Mean $\J$ & Mean $\F$ & Mean $\J\&\F$ \\ \midrule
	    & 		   & 72.0 & 73.6  & 72.8 \\ 
	    & \checkmark   & 73.2 & 75.0  & 74.1 \\ 
\checkmark  & 		   & 74.3 & 78.1  & 76.2 \\ 
\checkmark  & \checkmark   & \bf75.5  & \bf79.3  & \bf77.4 \\ \bottomrule
\end{tabular}}
\vspace{6pt}
\caption{Ablation study on online adaptation and spatio-temporal embedding}
\label{tab:ab_spatial_online_adapt}
\end{table}

\subsection{Interactive Video Object Segmentation}
Getting dense annotations in the first frame is a laborious and expensive process. It is therefore highly desirable that a system can interact with users in a more realistic way and reach the target quality with as little effort as possible. Our system allows users to interact with the system in real time, and see the result immediately after their input. In this section we consider the scenario of interactive video object segmentation, where the users are allowed to annotate any frame. The process is iterated and the user decides how to annotate based on the result up to the given point.

For the sake of simplicity, we limit the interaction to clicks: users can click the object of interest or the background. This way, the amount of interaction can easily be quantified as number of clicks. Please note though, that other types of interactions such as scribbles are also naturally supported by our system, although more difficult to evaluate in this experiment.

We first simulate the user behavior by a \textit{robot}. The robot randomly selects one pixel from the foreground and one pixel from the background as the first annotations, thus the nearest neighbor search can be performed. After having the initial result, the robot iteratively refines the segmentation result by randomly selecting from the pixels where the predicted label is wrong, and correcting its label based on the ground-truth. 

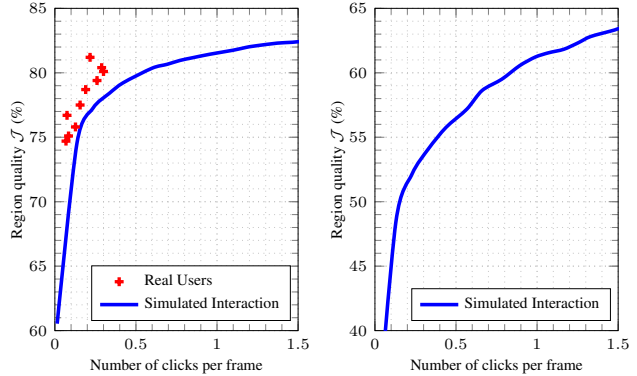
\begin{figure}[h]
\centering
\mbox{%
\resizebox{0.5\linewidth}{!}{\begin{tikzpicture}[/pgfplots/width=0.65\linewidth, /pgfplots/height=0.8\linewidth, /pgfplots/legend pos=south east]
    \begin{axis}[ymin=60,ymax=85,xmin=0,xmax=1.5,enlargelimits=false,
        xlabel=Number of clicks per frame,
        ylabel=Region quality $\J$ (\%),
		font=\scriptsize,
        grid=both,
		grid style=dotted,
        xlabel shift={-2pt},
        ylabel shift={-5pt},
        legend columns=1,
        minor ytick={60,61,...,85},
        ytick={60,65,...,85},
        minor xtick={0,0.1,...,2},
        xtick={0,0.5,...,2},
        legend pos= south east,
        legend cell align={left}
        ]
        \addplot[red,mark=+,only marks,line width=1.5] table [x=clicks, y=J]{data/user_click_J.txt};
        \addlegendentry{Real Users}
        \label{fig:real_users}
        \addplot[smooth,blue,mark=none, ultra thick] table [x=clicks, y=J]{data/robot_click_J.txt};
        \addlegendentry{Simulated Interaction}
        \label{fig:simulated_robot}
    \end{axis}
\end{tikzpicture}}
\resizebox{0.5\linewidth}{!}{\begin{tikzpicture}[/pgfplots/width=0.65\linewidth, /pgfplots/height=0.8\linewidth, /pgfplots/legend pos=south east]
    \begin{axis}[ymin=40,ymax=65,xmin=0,xmax=1.5,enlargelimits=false,
        xlabel=Number of clicks per frame,
        ylabel=Region quality $\J$ (\%),
		font=\scriptsize,
        grid=both,
		grid style=dotted,
        xlabel shift={-2pt},
        ylabel shift={-5pt},
        legend columns=1,
        minor ytick={45,46,...,65},
        ytick={40,45,...,65},
        minor xtick={0,0.1,...,2},
        xtick={0,0.5,...,2},
        legend pos= south east,
        legend cell align={left}
        ]
     \addplot[smooth,blue,mark=none, ultra thick] table [x=clicks, y=J]{data/robot_click_J_multiple.txt};
        \addlegendentry{Simulated Interaction}
     \end{axis}
\end{tikzpicture}}}
   \caption{\textbf{Interactive Segmentation Results}: Achieved quality with respect to the number of clicks provided in the single-object (left) on DAVIS 2016 and multiple-object (right) scenarios on DAVIS 2017.}
   \label{fig:res_interactive}
   \vspace{-2mm}
\end{figure}

The left side of Figure~\ref{fig:res_interactive} (\ref{fig:simulated_robot}) shows the evolution of the quality of the result as more clicks are provided. We achieve an overlap of $\J=80\%$ with only $0.55$ clicks per frame, and the performance goes up to $\J=83\%$ with $2$ clicks per frame. Our method achieves the same result as when providing the full mask on the first frame ($\J=75.5\%$) using only $0.15$ clicks per frame. Due to the randomness of our experiment, each experiment is repeated for $5$ times and we report the average overlap. We find the variance to be only $0.1$ at $1$ click per frame, which suggests that our method is reasonably robust to the selection of points. 

To verify that the simulated clicks are realistic, we carry out a user study on \textit{real} users, where we ask them to click freely until they are happy with the segmentation. The results are shown as points (\ref{fig:real_users}) in Figure~\ref{fig:res_interactive}. We can see that the real-user results are slightly better than the simulated ones, which we attribute to the fact that a real user can choose which point to click based on a global view (for instance, select the worst frame) instead of the random sampling that the robot performs. 

On average, the user did $0.17$ clicks per frame to achieve an overall result of $\J=77.7\%$. This equals to $11$ clicks per video, which takes around $24$ seconds. In contrast, a user takes 79 seconds to segment an object at the MS COCO quality~\cite{Lin2014}, so the full mask of the first frame at the quality of DAVIS can safely be estimated to take over 3 minutes. The quality achieved in these 24 seconds is comparable with most state-of-the-art semi-supervised methods, but at a fraction of the annotation and running cost.

\subsection{Extension to Multiple Objects}
As discussed in Section~\ref{sec:main_method}, our method can naturally extend to the segmentation of multiple objects. To validate the effectiveness of our method in such scenario, we carry out experiments on DAVIS 2017~\cite{Pont-Tuset_arXiv_2017}, where each video has multiple objects, usually interacting with and occluding each other. 

We summarize our results in the right side of Figure~\ref{fig:res_interactive}: our method generalizes well to multiple objects and the results are comparable with most state-of-the-art methods. For instance, OSVOS achieves $57\%$ in $\J$. We match their results by only $0.5$ clicks per frame, which leads to a fraction of the processing time of the former.

\section{Conclusions}
This work presents a conceptually simple yet highly effective method for video object segmentation. The problem is casted as a pixel-wise retrieval in an embedding space learned via a modification of the triplet loss specifically designed for video object segmentation. This way, the annotated pixels on the video (via scribbles, segmentation on the first mask, clicks, etc.) are the reference samples, and the rest of pixels are classified via a simple and fast nearest-neighbor approach. We obtain results comparable to the state of the art in the semi-supervised scenario, but significantly faster. Since the computed embedding vectors do not depend on the user input, the method is especially well suited for interactive segmentation: the response to the input feedback can be provided almost instantly. In this setup, we reach the same quality than in the semi-supervised case with only 0.15 clicks per frame. The method also naturally generalizes to the multiple objects scenario.

\paragraph{Acknowledgements}
This project is supported by armasuisse.
{\small
\bibliographystyle{ieee}
\bibliography{cvpr2018}
}

\end{document}